\documentclass{iet-ell}

\usepackage{amsmath}
\usepackage{subcaption}
\usepackage{multirow}
\usepackage{lipsum}
\usepackage{xcolor}

\setcopyright{open}
\ietVolume{00}
\ietYear{2021}
\ietDoi{ell.10001}

\received{DD MMMM YYYY}
\accepted{DD MMMM YYYY}

\begin{document}

\title{Joint prototype and coefficient prediction for 3D instance segmentation}

\author[af1,af2]{Remco Royen}
\orcid{0000-0001-9639-7775}
\author[af1,af2]{Leon Denis}
\orcid{0000-0001-7564-5549}
\author[af1,af2]{Adrian Munteanu}
\orcid{0000-0001-7290-0428}

\affil[af1]{Department of Electronics and Informatics, Vrije Universiteit Brussel, B-1050 Brussels, Belgium}
\affil[af2]{imec, Kapeldreef 75, B-3001 Leuven, Belgium}

\corresp{E-mail: remco.royen@vub.be}

\begin{abstract}
3D instance segmentation is crucial for applications demanding comprehensive 3D scene understanding. In this paper, we introduce a novel method that simultaneously learns coefficients and prototypes. Employing an overcomplete sampling strategy, our method produces an overcomplete set of instance predictions, from which the optimal ones are selected through a Non-Maximum Suppression (NMS) algorithm during inference. The obtained prototypes are visualizable and interpretable. Our method demonstrates superior performance on S3DIS-blocks, consistently outperforming existing methods in mRec and mPrec. Moreover, it operates 32.9\% faster than the state-of-the-art. Notably, with only 0.8\% of the total inference time, our method exhibits an over 20-fold reduction in the variance of inference time compared to existing methods. These attributes render our method well-suited for practical applications requiring both rapid inference and high reliability.
\end{abstract}

\maketitle

\section{Introduction}
Instance segmentation \textcolor{black}{is receiving} considerable attention in the field of scene understanding, particularly with the rise of AI. This task distinguishes itself from semantic segmentation~\citep{lu2020puconv, joukovsky2020multi, royen2023resscal3d} by not only assigning a semantic class to each pixel but also differentiating between objects for a deeper understanding. \textcolor{black}{While 2D instance segmentation has been well-explored~\citep{he2017mask, bolya2019yolact, wang2020solo, tian2020conditional, wang2020solov2, fang2021instances, ke2022mask, zhu2022sharpcontour}, our focus is on 3D instance segmentation. This domain presents additional challenges due to the unstructured and unordered nature and the potential high variations in instance sizes of the point clouds.}

Existing methods in this domain can be broadly classified into two categories. Proposal-based methods~\citep{yang2019learning, yi2019gspn, liu2020learning, engelmann20203d, sun2023neuralbf} involve 3D object detection followed by semantic segmentation. A notable drawback of these methods lies in the criticality of proposal accuracy, as errors in the two consecutive tasks tend to accumulate. Clustering-based methods~\textcolor{black}{\citep{wang2018sgpn, wang2019associatively, lahoud20193d, pham2019jsis3d, elich20193d,  jiang2020pointgroup, han2020occuseg, he2020instance, zhao2020jsnet, he2021dyco3d, chen2021hierarchical, zhang2021point, chen2022jspnet}} compute discriminative features which are employed for object clustering. Some also compute additional features such as center offsets to aid the clustering~\textcolor{black}{\citep{lahoud20193d, jiang2020pointgroup, he2021dyco3d, chen2021hierarchical}}. A recent advancement, Mask3D~\citep{schult2023mask3d}, sidesteps the need for precise proposals or time-consuming clustering by refining initial low-level queries using transformers.

\begin{figure}[t]
    \centering
    \includegraphics[width=0.99\linewidth]{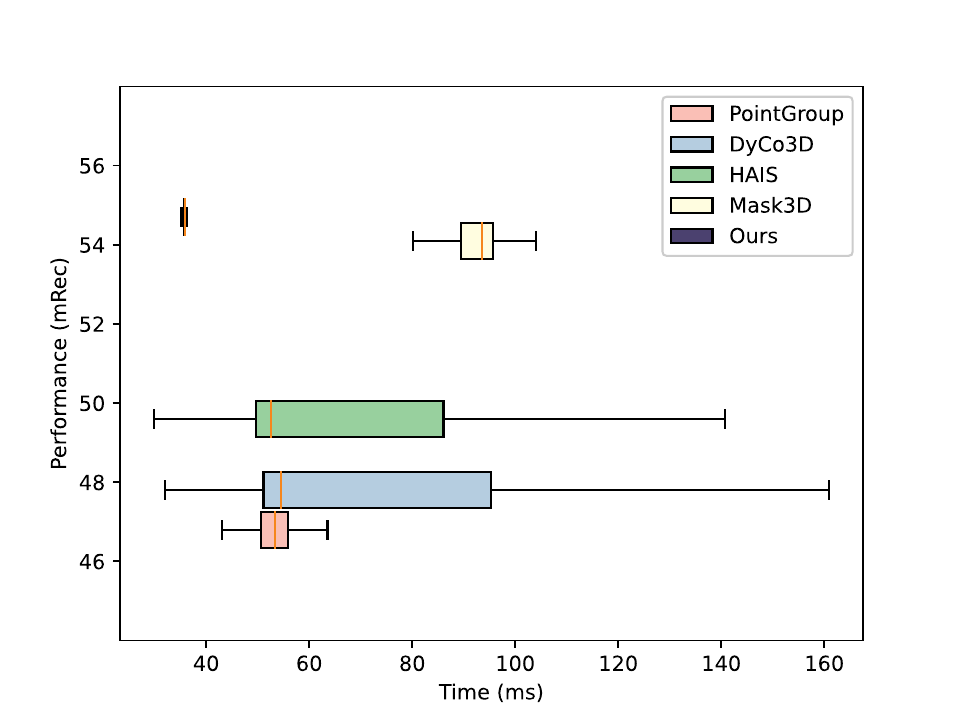}
    \caption{Speed-performance comparison on S3DIS-blocks Area-5. The proposed method outperforms the state-of-the-art in terms of accuracy, speed and variance in inference time.} 
    \label{fig:boxplot_speed_performance}
\end{figure}

In this paper, we introduce a novel 3D instance segmentation method that overcomes the mentioned drawbacks. Furthermore, it omits the time-expensive query refinement steps of Mask3D by jointly learning prototypes and coefficients. By employing a linear combination of these, we generate a set of instance predictions of which the best are retained through a rapid non-maximum suppression (NMS) algorithm. \textcolor{black}{A key aspect of our method is the overcompleteness of the obtained set of coefficients which are derived from a diverse set, with fixed cardinality, of sampled points in the input. This avoids the need for precise proposal prediction.} Not only does our approach outperform existing methods, it also demonstrates 32.9\% faster processing and exhibits a 20 times smaller variance in inference timings in comparison to existing methods, ensuring reliability and suitability for online applications. A performance and speed comparison with the state-of-the-art is visualized with a boxplot in Figure \ref{fig:boxplot_speed_performance}. In summary, our main contributions can be outlined as follows:


\begin{itemize}
    \item A novel, end-to-end prototype-based 3D instance segmentation method is proposed. Coefficients and prototypes are learned jointly and combined to predict the instance masks.
    \item A thorough validation on S3DIS blocks, showing state-of-the-art performance both on 5th-fold and 6-fold cross-validation (CV) settings. Additionally, the proposed method does not only achieve the fastest inference but also the lowest variance in inference time.
\end{itemize}

\begin{figure*}[t]
    \centering
    \includegraphics[width=0.99\linewidth]{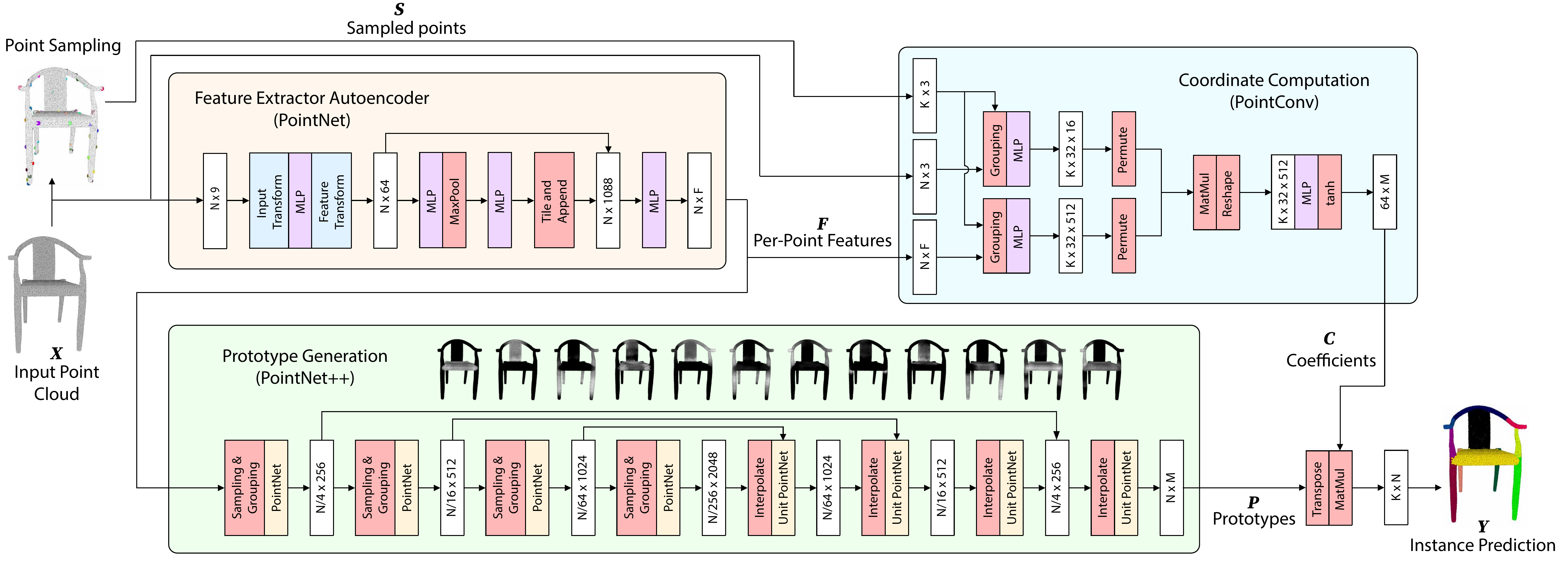}
    \caption{The proposed architecture consists of four main parts: (1) A feature extractor which retrieves per-point features. (2) The sampling of a diverse set of $K$ points (3) A PointNet++ network that generates prototypes from the per-point features (4) In parallel, a PointConv network that computes coefficients for each sampled point. Instance predictions are obtained by linearly combining the coefficients and prototypes from which the optimal ones are selected during inference.}
    \label{fig:arch}
\end{figure*}

\section{Proposed method}
The proposed method's architecture is illustrated in Figure \ref{fig:arch}. \textcolor{black}{Let $\boldsymbol{X} \in \mathbb{R}^{N \times I}$ be the input point cloud, where $N$ and $I$ represent the number of input points and channels, respectively. Per-point features $\boldsymbol{F} \in \mathbb{R}^{N \times F}$ for $\boldsymbol{X}$ are computed, with $F$ denoting the feature dimension.} These features are utilized in two parallel branches. The first branch samples a set $\boldsymbol{S} = \{s_{i}\}_{i=1}^{K}$ of $K$ points $s_i \in \mathbb{R}^{3}$ from the input data, \textcolor{black}{and} transforms the features based on the sampling into a coefficient set $\boldsymbol{C} \in \mathbb{R}^{K \times M}$, where $M$ is the number of prototypes. Simultaneously, the second branch computes $M$ prototypes, $\boldsymbol{P} \in \mathbb{R}^{N \times M}$, from $\boldsymbol{F}$. The linear combination of coefficients and prototypes enables the retrieval of the overcomplete set of instance masks $\boldsymbol{Y} \in \mathbb{R}^{K \times N}$. Mathematically, the combination of the coefficients and prototypes can be expressed as follows,
\begin{align}
\boldsymbol{y}_{k}^{q} = \sum_{i=1}^{M} c_{k, i}^{q} \boldsymbol{p}_{i}^{q},
\end{align}
with $\boldsymbol{y}_{k}^{q} = \boldsymbol{y}_{k}(\boldsymbol{X}_{q}) \in \mathbb{R}^{N}$ the $k^{th}$ prediction mask in $\boldsymbol{Y}$ for input sample $\boldsymbol{X}_{q}$; $c_{k,i}^{q} = c_{k,i}(\boldsymbol{X}_{q}) \in \mathbb{R}$ the $i^{th}$ coefficient for the k-th sampled point and $\boldsymbol{p}_{i}^{q} = \boldsymbol{p}_{i}(\boldsymbol{X}_{q}) \in \mathbb{R}^{N}$ the $i^{th}$ prototype for input sample $\boldsymbol{X}_{q}$.



During inference, $\boldsymbol{Y}$ is thresholded, and the $K'$ optimal predictions are selected using a non-maximum suppression (NMS) algorithm, resulting in the final instance masks $\boldsymbol{Y'} \in \mathbb{R}^{K' \times N}$. Subsequently, we will delve into the details of the coefficients and prototypes prediction.

\subsection{Coefficients} The coefficient computation begins with the sampling of $K$ points from the input point cloud using Farthest Point Sampling (FPS). Unlike traditional proposal-based methods, our approach doesn't necessitate precise proposal predictions but aims to acquire an overcomplete set of points showcasing large diversity in local neighborhoods. A PointConv network ~\citep{wu2019pointconv} calculates coefficients for each sampled point based on the features within their local neighborhood, contributing to a diverse set of coefficients. After a linear combination with prototypes, this yields a great variety in instance masks from which the NMS algorithm selects the optimal predictions. The PointConv network uses $tanh$ as its output activation function, accommodating negative coefficients and enabling the cancellation of specific regions among prototypes.

\subsection{Prototypes} 
\begin{figure}
  \centering
   \includegraphics[width=0.95\linewidth]{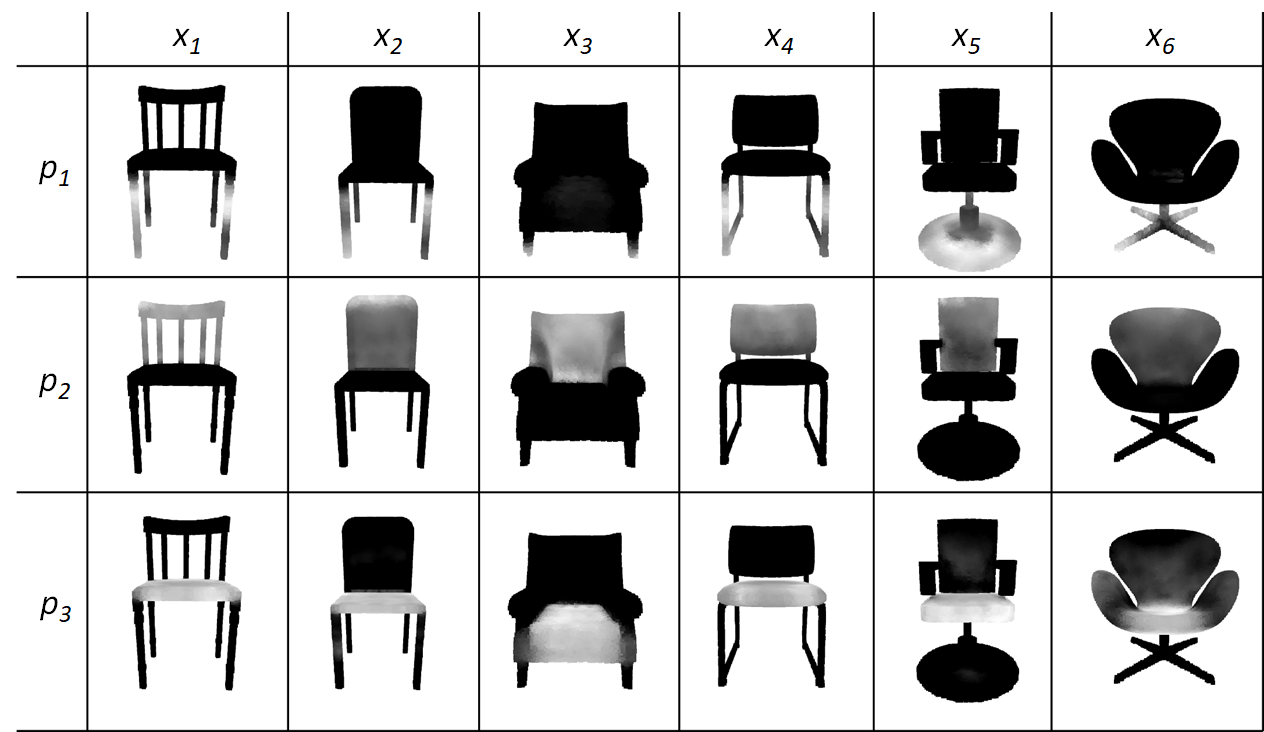}
   \caption{Prototype $p_i$ activation for different PartNet samples $x_i$.}
   \label{fig:proto_vis}
\end{figure}

A PointNet++ network~\citep{qi2017pointnet++} utilizes the shared features $\boldsymbol{F}$ to compute a set of prototypes $\boldsymbol{P} \in \mathbb{R}^{N \times M}$, consisting of $M$ prototypes $\boldsymbol{p_{i}} \in \mathbb{R}^{N}$. It is noteworthy that these prototypes maintain the point dimension of the input, effectively forming vectors consisting of scores which are associated to each input point. Visual representations of prototype scores on input point clouds from the PartNet-chair dataset~\citep{mo2019partnet} are presented in Figure \ref{fig:proto_vis}. In these visualizations, white and black points signify high and low scores in a given prototype, respectively. Notably, each prototype, i.e. row in Figure \ref{fig:proto_vis}, emphasizes a specific region of the input point cloud, encapsulating instance information. The linear combinations of such prototypes enable the generation of instance masks.

\subsection{Loss} The proposed method is trained end-to-end. As loss-function, we employ a binary cross entropy loss between the instance masks $\boldsymbol{Y_{i}}$ and the ground-truth instance masks $\boldsymbol{GT_{i}}$. To avoid instability and accelerate training, we introduce a spatial constraint where the predicted instance mask associated to the sampled point $s_{i}$, $\boldsymbol{Y}(s_{i})$\textcolor{black}{,} should match the ground-truth instance that contains the sampled point, $\boldsymbol{GT}(s_{i}))$. Mathematically, this is expressed as follows:
\begin{align}
  J = \sum_{i=1}^{I_{PR}} L_{BCE}(\boldsymbol{Y}(s_{i}), \boldsymbol{GT}(s_{i})), 
\end{align}
with $I_{PR}$ the number of predicted instances.
\subsection{Experiments}
To assess the efficacy of our proposed method, we conduct qualitative and quantitative experiments on the S3DIS dataset~\citep{armeni20163d}, encompassing 6 large-scale indoor areas with a total of 271 rooms. Each point in the dataset is annotated with one of 13 semantic categories. Our evaluation strategy and metrics align with established practices in the literature~\citep{zhao2020jsnet}.

\begin{figure}
\centering
\includegraphics[width=0.79\linewidth]{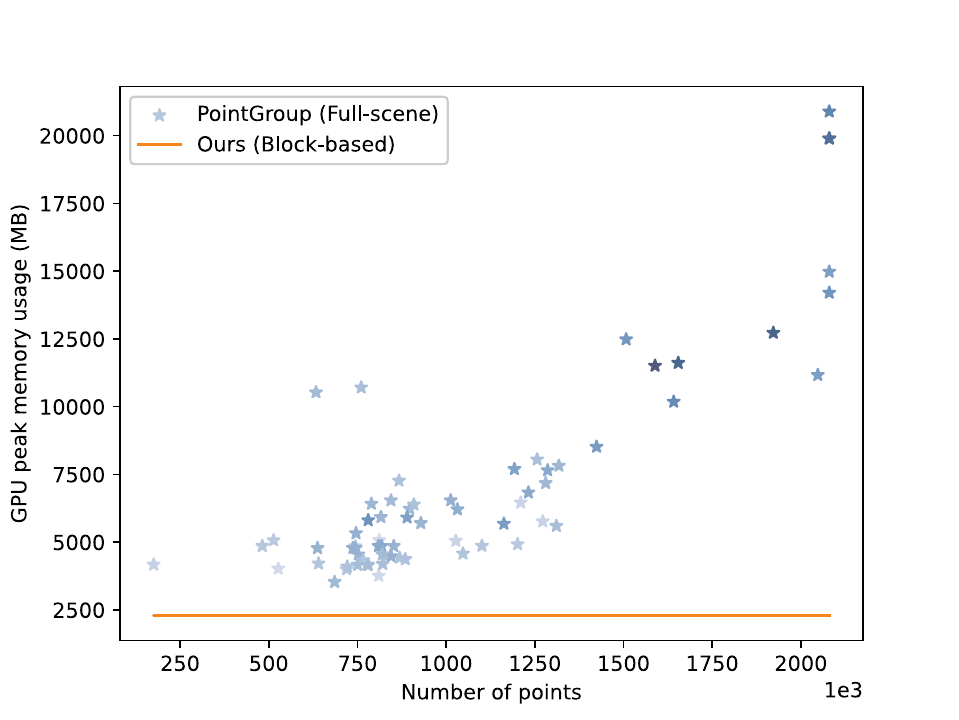}
\caption{Peak memory usage of the proposed method (block-based) and PointGroup \citep{jiang2020pointgroup} (full-scene) for different number of input points and number of instances (intensity of marker). The experiment is conducted on the scenes of S3DIS Area-5 \citep{armeni20163d}}
\label{fig:mem_reqs}
\end{figure}

In the literature, two primary approaches for handling large-scale and dense datasets such as S3DIS are recognized. \textcolor{black}{One, utilized by \citep{wang2018sgpn, zhao2020jsnet, he2020learning, he2020instance, denis2023improved}, firstly segments scenes into overlapping blocks. These are processed separately to instance masks, and merged using the BlockMerge algorithm~\citep{wang2018sgpn, denis2023improved}.} The other, found in \citep{jiang2020pointgroup, he2021dyco3d, chen2021hierarchical, vu2022softgroup}, processes the entire scene at once without merging. While the latter offers a higher segmentation performance as it avoids partial objects and provides a global overview, it incurs significant memory requirements, as depicted in Figure \ref{fig:mem_reqs}. PointGroup, favoring whole-scene processing, experiences escalating memory demands with an increasing number of points, exceeding 20 GB for the largest scenes. Notably, even on our powerful RTX 3090 GPU, four scenes were unable to be processed as a whole, necessitating downsampling to obtain results. In contrast, block-based methods, like ours, maintain a consistent GPU memory requirement for processing point clouds of arbitrary sizes. As example, our method requires a constant 2.4 GB peak GPU memory. Block-based (BB) methods are also suitable for online applications where the scene has to be processed during capturing. While not reaching the same performance as full-scene methods, block-based approaches remain relevant for practical applications, particularly on embedded devices or in online scenarios. Due to computational constraints, ProtoSeg couldn't be applied to the entire scene, and we present our results solely using a block-based approach (S3DIS-blocks).

\begin{table*}[t]
\begin{center}
\caption{Instance segmentation result on S3DIS-blocks for both Area-5 and 6-fold cross validation and timing. The methods denoted with * were retrained on blocks. Instance segmentation inference time and standard deviation are reported for S3DIS-blocks Area-5. For fair comparison, all runtimes were measured on the same CPU and NVIDIA GeForce RTX 3090. The '-' symbol is employed for non-public codebases, impeding the measurement of inference time. The '/' symbol indicates that a method does not incorporate that specific step.}
\begin{tabular}{|c|c c c c | c c c c| c c c c|} 
    \cline{2-13}
    \multicolumn{1}{c|}{} & \multicolumn{4}{c|}{5th-fold} & \multicolumn{4}{c|}{6-fold CV} & \multicolumn{4}{c|}{Inference time (ms)}\\
    \hline
    Method & mCov & mWCov & mRec & mPrec & mCov & mWCov & mRec & mPrec & Network & NMS & Grouping & \multicolumn{1}{|c|}{Total}\\ 
    \hline
    JSNet~\citep{zhao2020jsnet}               & 48.7 & 51.5 & 46.9 & 62.1 & 54.1 & 58.0 & 53.9 & 66.9 & 34.0 $\pm$ 11.3 & / & 96.2 $\pm$ 70.4 & \multicolumn{1}{|c|}{133.6 $\pm$ 71.0}\\ 
    IAM~\citep{he2020instance}                & 49.9 & 53.2 & 48.5 & 61.3 & 54.5 & 58.0 & 51.8 & 67.2 & - & - & - & \multicolumn{1}{|c|}{-}\\ 
    MPNet~\citep{he2020learning}               & 50.1 & 53.2 & 49.0 & 62.5 & \bf{55.8} & 59.7 & 53.7 & 68.4 & - & - & - & \multicolumn{1}{|c|}{-}\\ 
    PointGroup*~\citep{jiang2020pointgroup}  & 43.1 & 47.8 & 46.8 & 55.1 & 48.4 & 54.4 & 55.3 & 59.4& 47.1 $\pm$ 6.6 & 0.4 $\pm$ 0 & 4.8 $\pm$ 1.5 & \multicolumn{1}{|c|}{53.2 $\pm$ 6.3}\\
    DyCo3D*~\citep{he2021dyco3d}             & 44.4 & 50.0 & 47.8 & 59.3 & 47.9 & 54.5 & 52.4 & 58.9& 63.3 $\pm$ 28.7 & 0.5 $\pm$ 0.7 & 6.5 $\pm$ 7.3 & \multicolumn{1}{|c|}{72.7 $\pm$ 35.4}\\
    HAIS*~\citep{chen2021hierarchical}        & 45.4 & 50.6 & 49.6 & 58.4 & 48.2 & 55.4 & 55.0 & 60.7& 57.4 $\pm$ 23.3 & / & 9.1 $\pm$ 10.0 & \multicolumn{1}{|c|}{69.3 $\pm$ 30.5}\\
    SoftGroup*~\citep{vu2022softgroup}        & 40.6 & 45.6 & 43.9 & 42.6 & 45.0 & 51.2 & 48.4 & 52.3& 43.8 $\pm$ 3.5 & / & 5.6 $\pm$ 2.7 & \multicolumn{1}{|c|}{55.4 $\pm$ 6.0}\\
    JSPNet~\citep{chen2022jspnet}            & 50.7 & 53.5 & 48.0 & 59.6 & 54.9 & 58.8 & 55.0 & 66.5 & - & - & - & \multicolumn{1}{|c|}{-}\\ 
    Mask3D*~\citep{schult2023mask3d}            & \bf{51.4} & 56.7 & 54.1 & 59.9 & 52.3 & \textbf{61.0} & 56.2 & 67.1 & 73.8 $\pm$ 4.7 & / & / & \multicolumn{1}{|c|}{92.5 $\pm$ 8.3}\\ 
    \hline
    Ours                                    & 50.8 & \bf{57.0} & \bf{54.7} & \bf{65.0} & 52,5 & 59,9 & \textbf{56,5} & \textbf{70,5} & 35.0 $\pm$ 0.3 & 0.4 $\pm$ 0 & / & \multicolumn{1}{|c|}{\bf{35.7 $\pm$ 0.3}}\\
    \hline
\end{tabular}
\label{tab:s3dis}
\end{center}
\end{table*}

\begin{figure*}[t]
    \centering
    \begin{subfigure}[b]{0.3\textwidth}
        \centering
        \includegraphics[width=\textwidth]{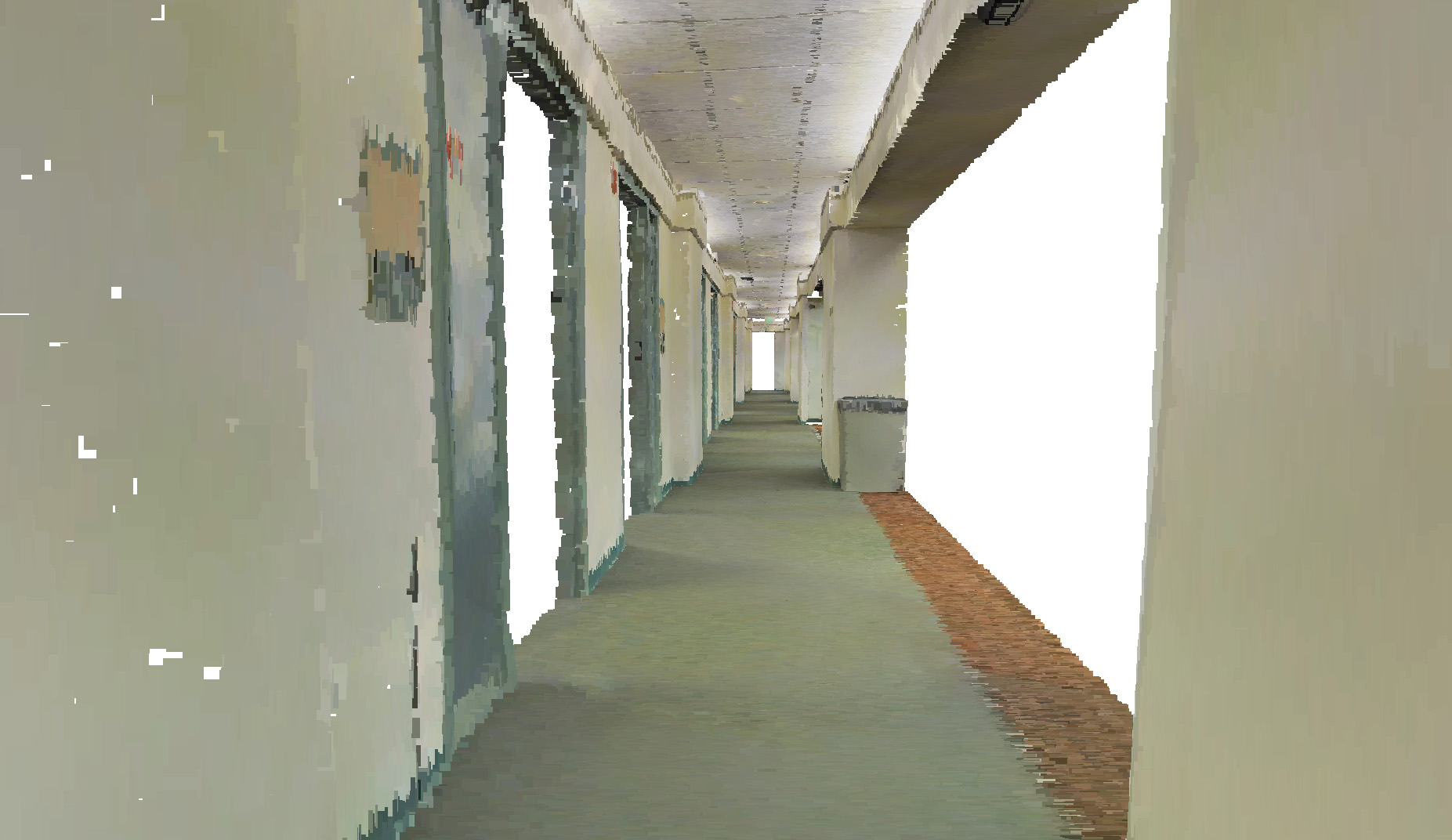}
    \end{subfigure}
    \begin{subfigure}[b]{0.3\textwidth}
        \centering
        \includegraphics[width=\textwidth]{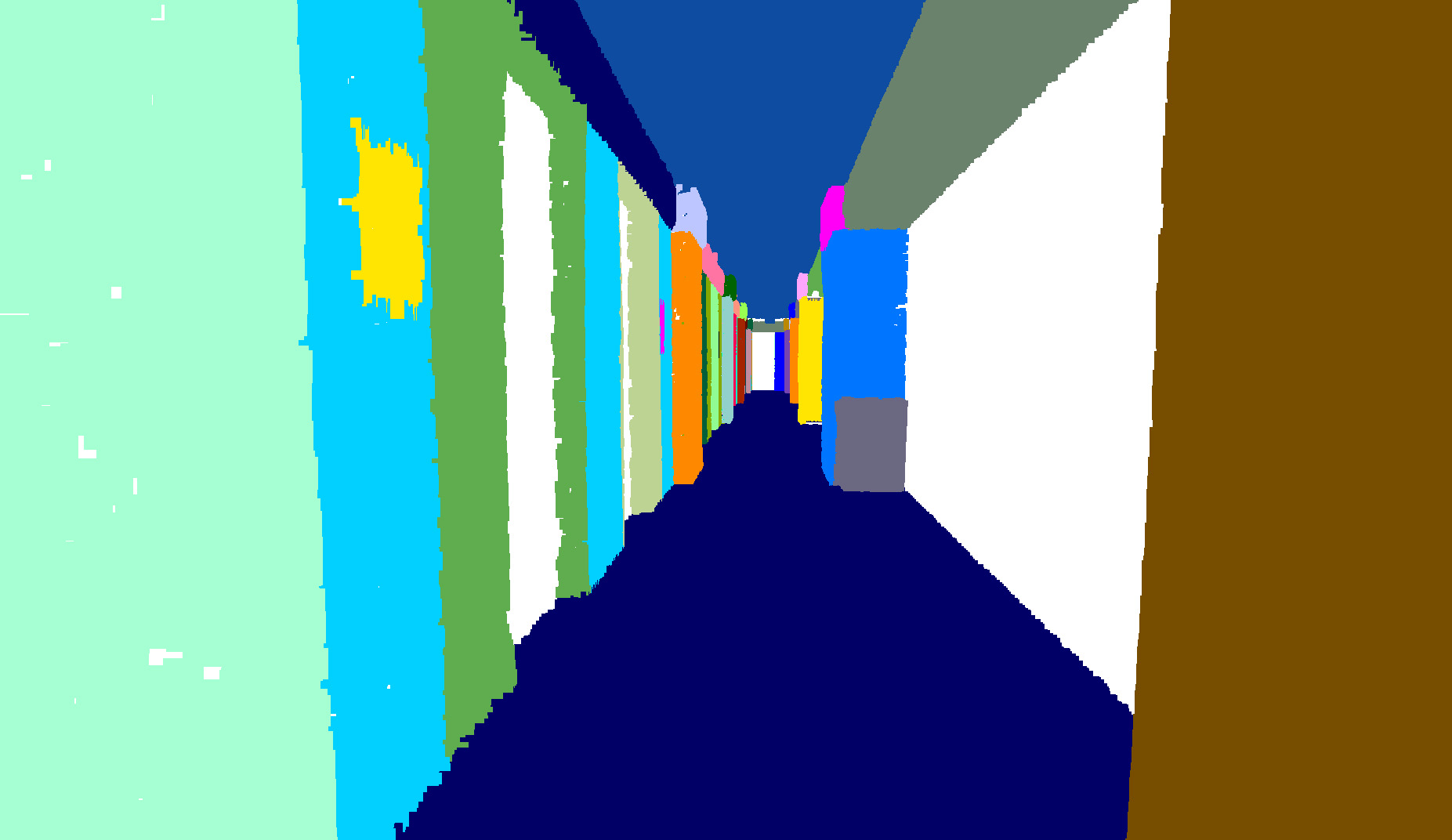}
    \end{subfigure}
    \begin{subfigure}[b]{0.3\textwidth}
        \centering
        \includegraphics[width=\textwidth]{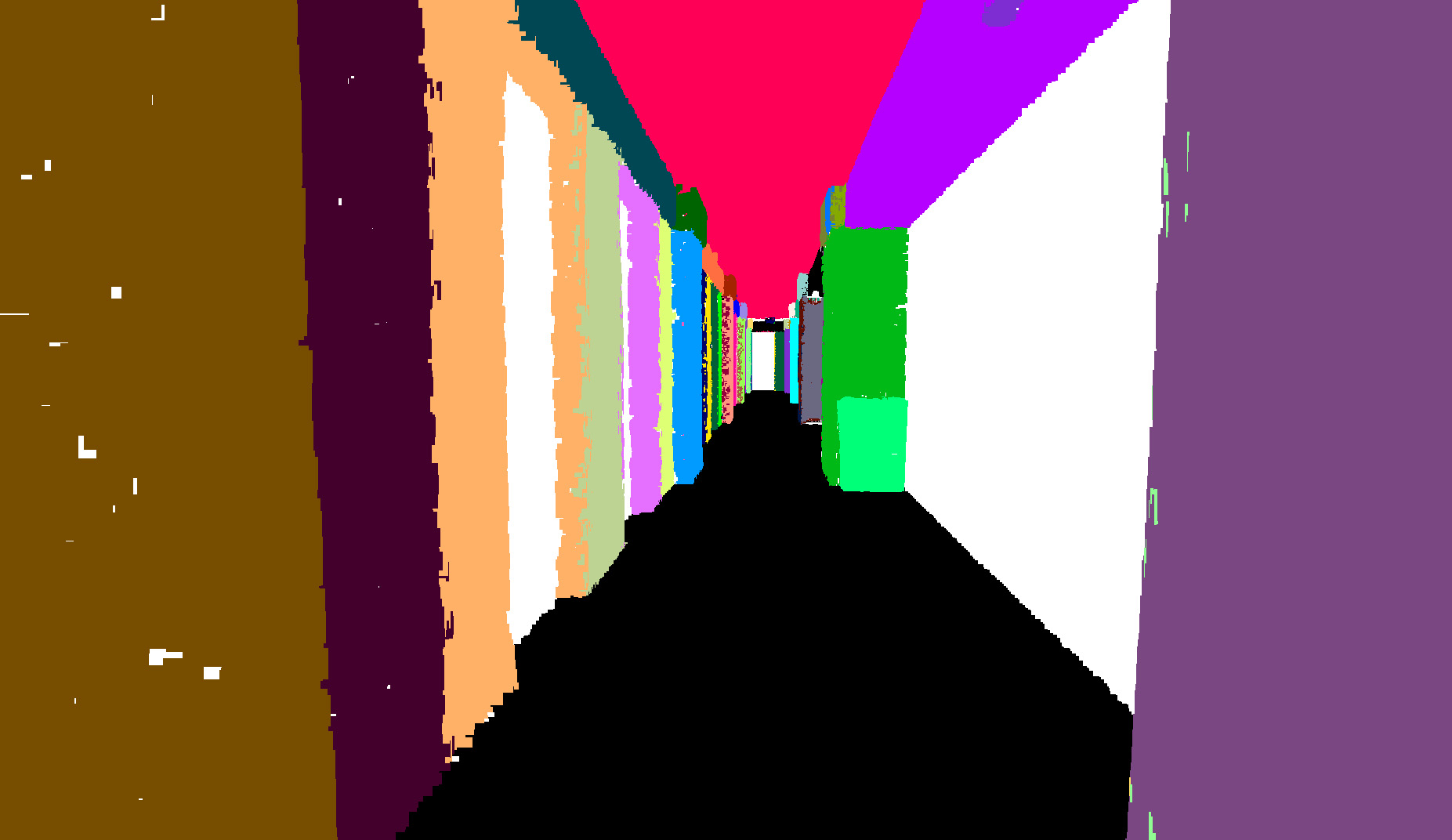}
    \end{subfigure}

\centering
    \begin{subfigure}[b]{0.3\textwidth}
        \centering
        \includegraphics[width=\textwidth]{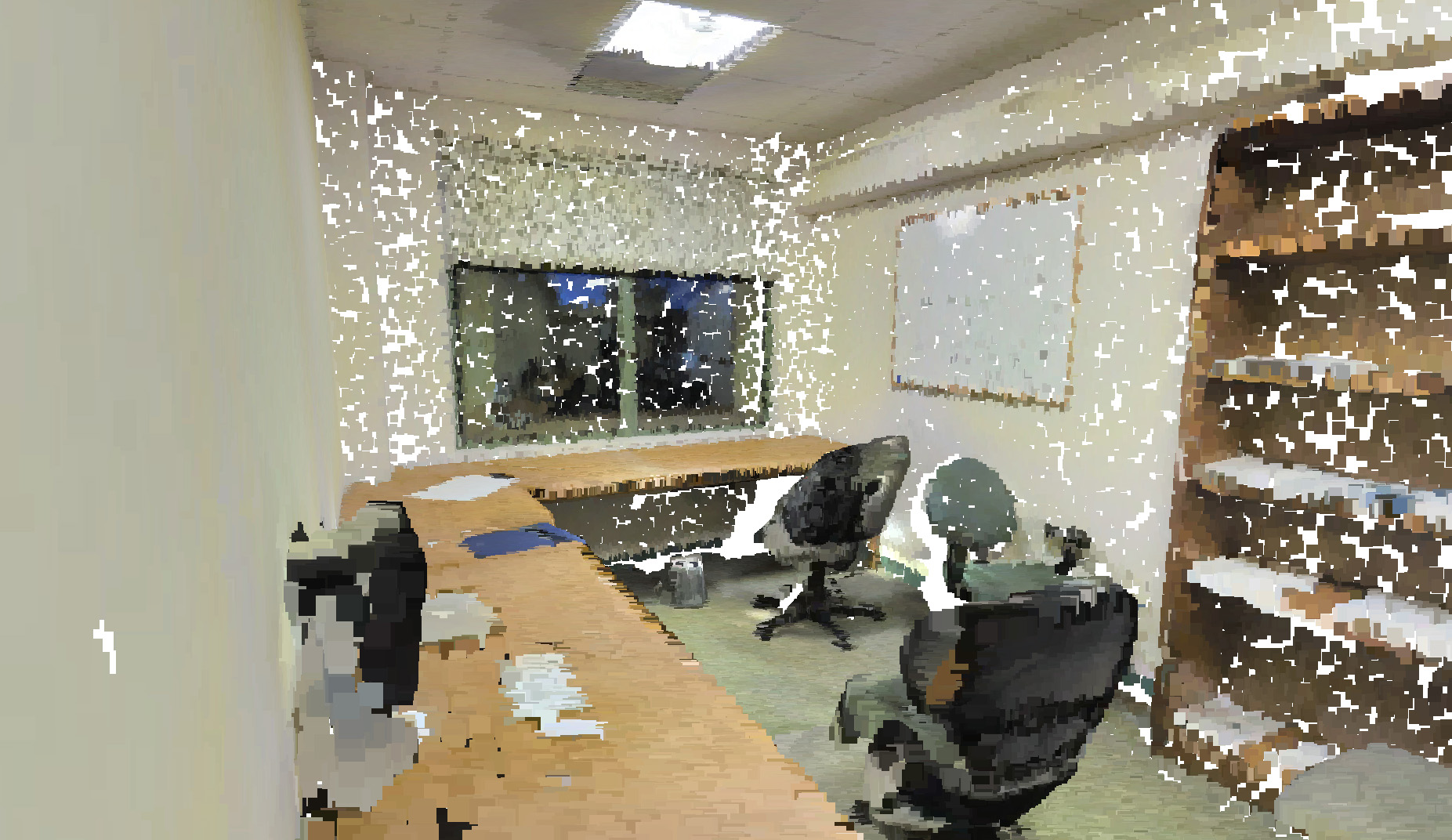}
        \caption{Input}
        \label{subfig:s3dis_vis_input}
    \end{subfigure}
    \begin{subfigure}[b]{0.3\textwidth}
        \centering
        \includegraphics[width=\textwidth]{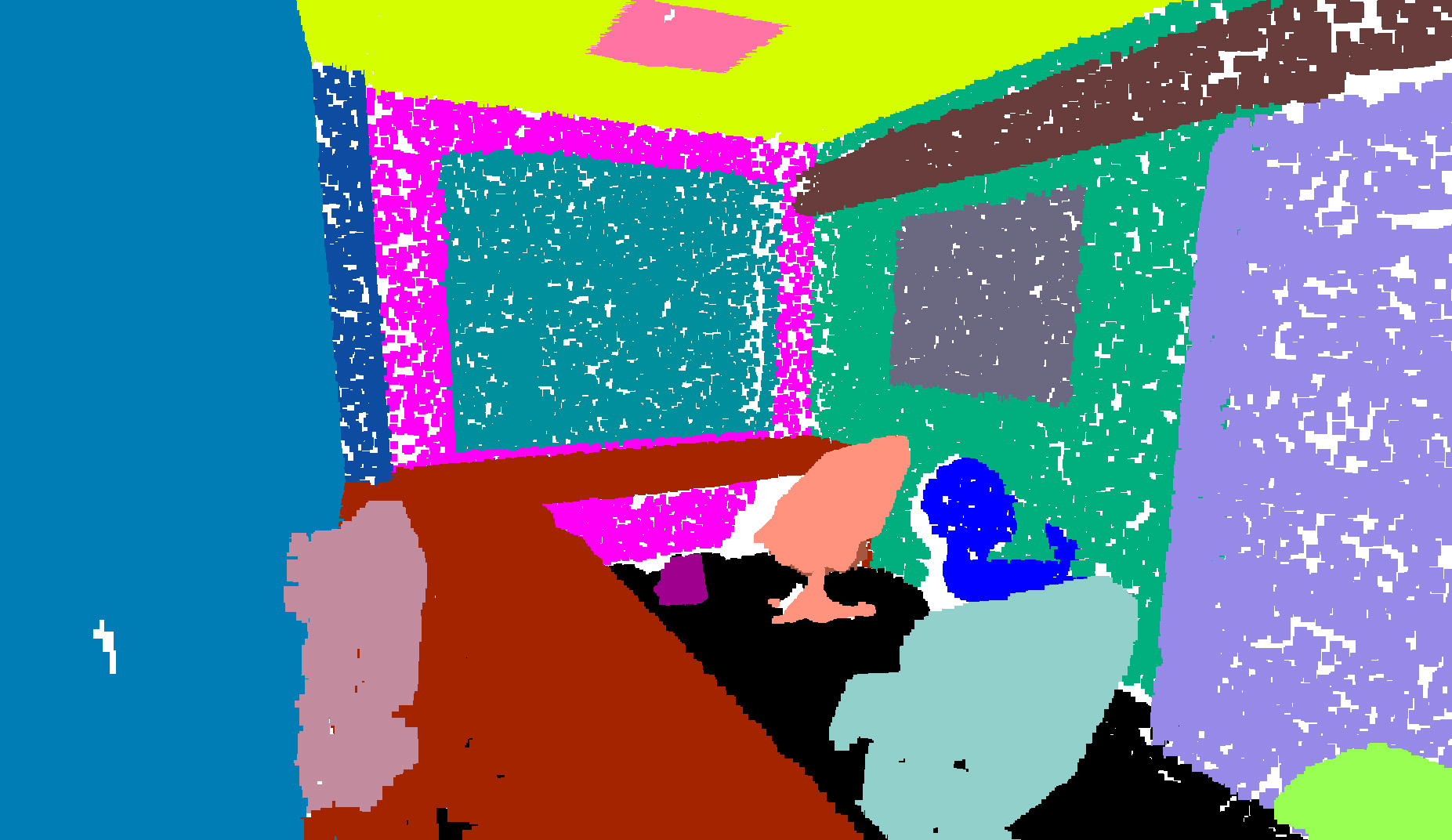}
        \caption{Instance GT}
        \label{subfig:s3dis_vis_gt}
    \end{subfigure}
    \begin{subfigure}[b]{0.3\textwidth}
        \centering
        \includegraphics[width=\textwidth]{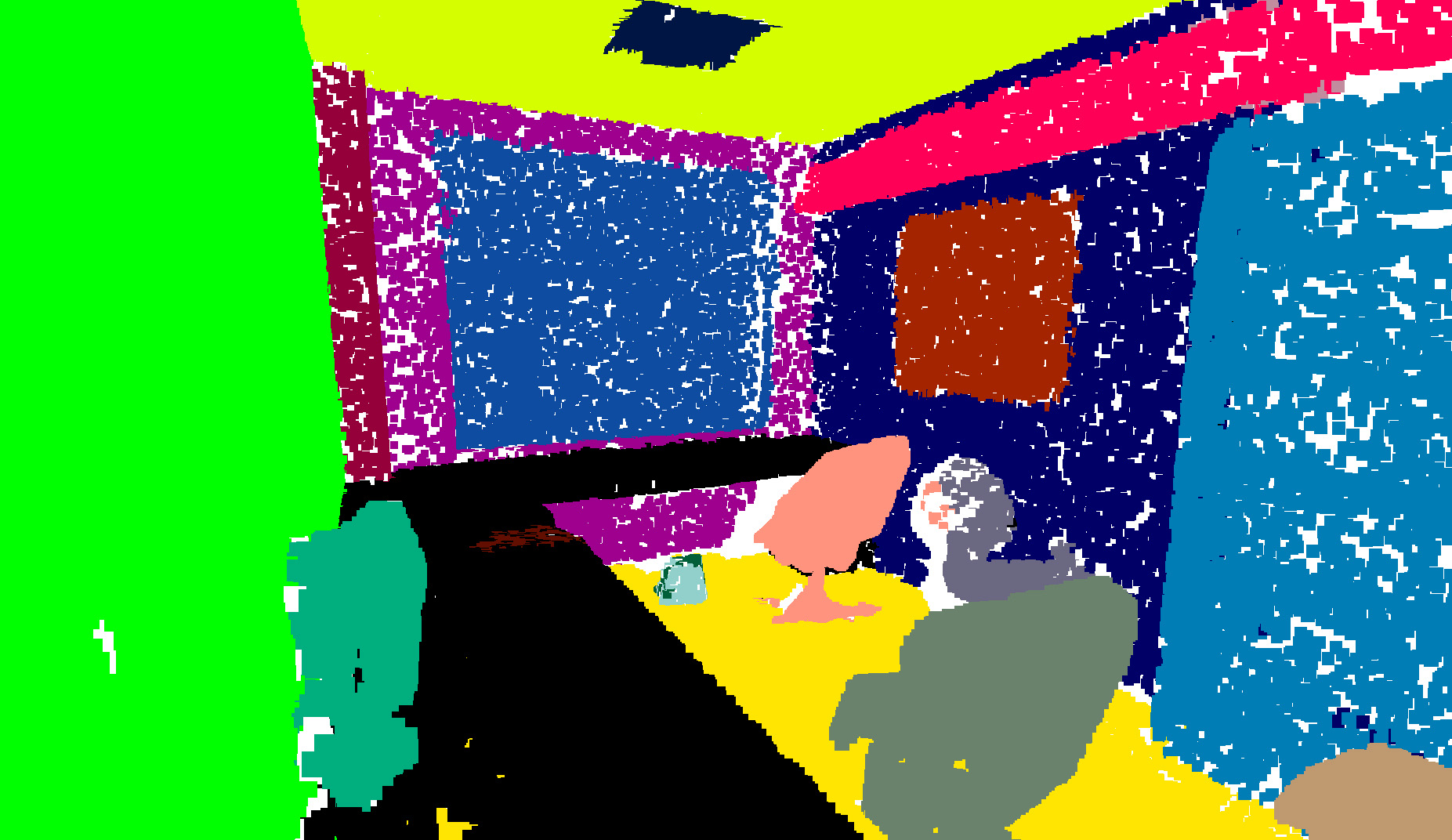}
        \caption{Instance predictions}
        \label{subfig:s3dis_vis_pred}
    \end{subfigure}
    
    \caption{Visualization of S3DIS results. Note that different colors represent different instances and that the same instance may have a different color in the ground-truth and prediction}
    \label{fig:s3dis_vis}
\end{figure*}

We follow the experimental settings of \citep{qi2017pointnet, zhao2020jsnet} and divide each room in 1m $\times$ 1m blocks with stride 0.5 such that an overlap is created. In each block 4096 points are randomly sampled during training. Each input point is represented by a 9-dimensional vector: XYZ, RGB and the normalized location of the point in the whole room. We have used a PointNet \citep{qi2017pointnet} feature extractor, a PointConv without densitynet \citep{wu2019pointconv} to compute the coefficients and PointNet++ \citep{qi2017pointnet++} for the prototypes. Semantic segmentation is obtained, in a block-based manner, by a PointTransformer \citep{zhao2021point} network. An Adam optimizer \citep{kingma2014adam} is used with learning rate 0.001 and the network is trained with batch size 16 for 65 epochs. We have empirically selected the number of features, prototypes, sampled points and the sampling method as 64, 128, 64 and 'FPS on xyz', respectively, as it led to the best results. During inference, we employ a threshold of 0.3 and use a regular NMS to retrieve the instance mask predictions for each block. The same BlockMerge algorithm as proposed in \citep{wang2018sgpn} is employed to ensure a fair comparison between block-based methods.

\subsection{Experimental results on S3DIS-blocks}
Quantitative results on S3DIS-blocks, along with comparisons with existing methods, are presented in Table \ref{tab:s3dis}. Note that for a fair comparison, we retrained methods that do not operate on blocks using their official, publicly available code. 

On the Area-5 validation, our method surpasses existing works in three out of the four metrics and secures the second-best result in the remaining one. More specifically, we achieve a 0.6\% higher mean recall (mRec) and a notable 2.5\% increase in mean precision (mPrec) over the closest competitor. Similar trends are observed in the 6-fold cross-validation, where our method slightly outperforms Mask3D by 0.3\% in mRec and significantly surpasses the second-best by 2.1\% in mPrec. Regarding timing, the proposed method stands out with a total inference time of 35.7 ms, making it 32.9\% faster than the current fastest method. Additionally, our overcomplete approach significantly reduces the variance in inference time to just 0.8\% of our total inference time, or 0.3 ms. This is over 20 times smaller than the smallest variance observed in existing methods.

Qualitative results are showcased in Figure \ref{fig:s3dis_vis}. Overall, one can observe highly accurate segmentation, with nearly all instances detected in the corridor. Similarly, for the office, the segmentation results are notably accurate, encompassing the detection of chairs, walls, and apparel.

\section{Conclusion}
In conclusion, this paper presents a novel 3D instance segmentation method designed for accurate, fast and reliable scene understanding. The proposed approach leverages an overcomplete sampling strategy and simultaneously learns coefficients and prototypes. Through experimentation on the S3DIS-blocks dataset, our method demonstrates superior performance, consistently outperforming existing methods in key metrics such as mean recall and mean precision. Notably, with a total inference time of 35.7 ms, our method is 32.9\% faster than the current fastest method. An additional strength of our method lies in its ultra-low variance in inference time, ensuring exceptional consistency and reliability.

In comparison with existing methodologies, our approach stands out as a promising solution for real-world applications demanding both accuracy and computational efficiency. We believe our contribution paves the way for further advancements in 3D instance segmentation, particularly in practical scenarios such as embedded devices or online applications.

\begin{acks}
This work is funded by Fonds Wetenschappelijk Onderzoek (FWO) - 1S89420N and Innoviris within the research project SPECTRE.

\noindent\textcolor{black}{\textit{Conflict of interest statement:} The authors declare no conflict of interest.}

\noindent\textcolor{black}{\textit{Data availability statement:} The data that support the findings of this study are available from the corresponding author upon reasonable request.}
\end{acks}

 \bibliographystyle{iet} 
 \bibliography{iet-ell}

\begin{thebibliography}{10}
\ifx\bibinfo\relax\providecommand{\bibinfo}[2]{#2}\fi
\makeatletter
\ifx\xfnm\@undefined\gdef\xfnm[#1]{#1}\fi
\ifx\xsnm\@undefined\gdef\xsnm[#1]{#1}\fi
\ifx\plxcitation\@undefined\def\plxcitation#1#2#3#4#5{}\fi
\ifx\endplxcitation\@undefined\def\endplxcitation{}\fi
\makeatother

\bibitem{lu2020puconv}
\plxcitation{}{lu2020puconv}{}{}{article}
\bibinfo{author}{\xsnm[Lu]\xfnm[, J.]}, et~al.: \bibinfo{title}{Puconv: Upsampling convolutional network for point cloud semantic segmentation}.
\newblock \bibinfo{journal}{Electronics letters} \bibinfo{volume}{56}(\bibinfo{number}{9}), \bibinfo{pages}{435--438} (\bibinfo{year}{2020})
\endplxcitation

\bibitem{joukovsky2020multi}
\plxcitation{}{joukovsky2020multi}{}{}{article}
\bibinfo{author}{\xsnm[Joukovsky]\xfnm[, B.]}, \bibinfo{author}{\xsnm[Hu]\xfnm[, P.]}, \bibinfo{author}{\xsnm[Munteanu]\xfnm[, A.]}: \bibinfo{title}{Multi-modal deep network for rgb-d segmentation of clothes}.
\newblock \bibinfo{journal}{Electronics Letters} \bibinfo{volume}{56}(\bibinfo{number}{9}), \bibinfo{pages}{432--435} (\bibinfo{year}{2020})
\endplxcitation

\bibitem{royen2023resscal3d}
\plxcitation{}{royen2023resscal3d}{}{}{inproceedings}
\bibinfo{author}{\xsnm[Royen]\xfnm[, R.]}, \bibinfo{author}{\xsnm[Munteanu]\xfnm[, A.]}: \bibinfo{title}{Resscal3d: Resolution scalable 3d semantic segmentation of point clouds}.
\newblock In: \bibinfo{booktitle}{2023 IEEE International Conference on Image Processing (ICIP)}, pp.~ \bibinfo{pages}{2775--2779}. \bibinfo{organization}{IEEE} (\bibinfo{year}{2023})
\endplxcitation

\bibitem{he2017mask}
\plxcitation{}{he2017mask}{}{}{inproceedings}
\bibinfo{author}{\xsnm[He]\xfnm[, K.]}, et~al.: \bibinfo{title}{Mask r-cnn}.
\newblock In: \bibinfo{booktitle}{Proceedings of the IEEE international conference on computer vision}, pp.~ \bibinfo{pages}{2961--2969}.  (\bibinfo{year}{2017})
\endplxcitation

\bibitem{bolya2019yolact}
\plxcitation{}{bolya2019yolact}{}{}{inproceedings}
\bibinfo{author}{\xsnm[Bolya]\xfnm[, D.]}, et~al.: \bibinfo{title}{Yolact: Real-time instance segmentation}.
\newblock In: \bibinfo{booktitle}{Proceedings of the IEEE/CVF international conference on computer vision}, pp.~ \bibinfo{pages}{9157--9166}.  (\bibinfo{year}{2019})
\endplxcitation

\bibitem{wang2020solo}
\plxcitation{}{wang2020solo}{}{}{inproceedings}
\bibinfo{author}{\xsnm[Wang]\xfnm[, X.]}, et~al.: \bibinfo{title}{Solo: Segmenting objects by locations}.
\newblock In: \bibinfo{booktitle}{European Conference on Computer Vision}, pp.~ \bibinfo{pages}{649--665}. \bibinfo{organization}{Springer} (\bibinfo{year}{2020})
\endplxcitation

\bibitem{tian2020conditional}
\plxcitation{}{tian2020conditional}{}{}{inproceedings}
\bibinfo{author}{\xsnm[Tian]\xfnm[, Z.]}, \bibinfo{author}{\xsnm[Shen]\xfnm[, C.]}, \bibinfo{author}{\xsnm[Chen]\xfnm[, H.]}: \bibinfo{title}{Conditional convolutions for instance segmentation}.
\newblock In: \bibinfo{booktitle}{European Conference on Computer Vision}, pp.~ \bibinfo{pages}{282--298}. \bibinfo{organization}{Springer} (\bibinfo{year}{2020})
\endplxcitation

\bibitem{wang2020solov2}
\plxcitation{}{wang2020solov2}{}{}{article}
\bibinfo{author}{\xsnm[Wang]\xfnm[, X.]}, et~al.: \bibinfo{title}{Solov2: Dynamic and fast instance segmentation}.
\newblock \bibinfo{journal}{Advances in Neural information processing systems} \bibinfo{volume}{33}, \bibinfo{pages}{17721--17732} (\bibinfo{year}{2020})
\endplxcitation

\bibitem{fang2021instances}
\plxcitation{}{fang2021instances}{}{}{inproceedings}
\bibinfo{author}{\xsnm[Fang]\xfnm[, Y.]}, et~al.: \bibinfo{title}{Instances as queries}.
\newblock In: \bibinfo{booktitle}{Proceedings of the IEEE/CVF International Conference on Computer Vision}, pp.~ \bibinfo{pages}{6910--6919}.  (\bibinfo{year}{2021})
\endplxcitation

\bibitem{ke2022mask}
\plxcitation{}{ke2022mask}{}{}{inproceedings}
\bibinfo{author}{\xsnm[Ke]\xfnm[, L.]}, et~al.: \bibinfo{title}{Mask transfiner for high-quality instance segmentation}.
\newblock In: \bibinfo{booktitle}{Proceedings of the IEEE/CVF Conference on Computer Vision and Pattern Recognition}, pp.~ \bibinfo{pages}{4412--4421}.  (\bibinfo{year}{2022})
\endplxcitation

\bibitem{zhu2022sharpcontour}
\plxcitation{}{zhu2022sharpcontour}{}{}{inproceedings}
\bibinfo{author}{\xsnm[Zhu]\xfnm[, C.]}, et~al.: \bibinfo{title}{Sharpcontour: A contour-based boundary refinement approach for efficient and accurate instance segmentation}.
\newblock In: \bibinfo{booktitle}{Proceedings of the IEEE/CVF Conference on Computer Vision and Pattern Recognition}, pp.~ \bibinfo{pages}{4392--4401}.  (\bibinfo{year}{2022})
\endplxcitation

\bibitem{yang2019learning}
\plxcitation{}{yang2019learning}{}{}{article}
\bibinfo{author}{\xsnm[Yang]\xfnm[, B.]}, et~al.: \bibinfo{title}{Learning object bounding boxes for 3d instance segmentation on point clouds}.
\newblock \bibinfo{journal}{Advances in neural information processing systems} \bibinfo{volume}{32} (\bibinfo{year}{2019})
\endplxcitation

\bibitem{yi2019gspn}
\plxcitation{}{yi2019gspn}{}{}{inproceedings}
\bibinfo{author}{\xsnm[Yi]\xfnm[, L.]}, et~al.: \bibinfo{title}{Gspn: Generative shape proposal network for 3d instance segmentation in point cloud}.
\newblock In: \bibinfo{booktitle}{Proceedings of the IEEE/CVF Conference on Computer Vision and Pattern Recognition}, pp.~ \bibinfo{pages}{3947--3956}.  (\bibinfo{year}{2019})
\endplxcitation

\bibitem{liu2020learning}
\plxcitation{}{liu2020learning}{}{}{article}
\bibinfo{author}{\xsnm[Liu]\xfnm[, S.H.]}, et~al.: \bibinfo{title}{Learning gaussian instance segmentation in point clouds}.
\newblock \bibinfo{journal}{arXiv preprint arXiv:2007.09860}  (\bibinfo{year}{2020})
\endplxcitation

\bibitem{engelmann20203d}
\plxcitation{}{engelmann20203d}{}{}{inproceedings}
\bibinfo{author}{\xsnm[Engelmann]\xfnm[, F.]}, et~al.: \bibinfo{title}{3d-mpa: Multi-proposal aggregation for 3d semantic instance segmentation}.
\newblock In: \bibinfo{booktitle}{Proceedings of the IEEE/CVF conference on computer vision and pattern recognition}, pp.~ \bibinfo{pages}{9031--9040}.  (\bibinfo{year}{2020})
\endplxcitation

\bibitem{sun2023neuralbf}
\plxcitation{}{sun2023neuralbf}{}{}{inproceedings}
\bibinfo{author}{\xsnm[Sun]\xfnm[, W.]}, et~al.: \bibinfo{title}{Neuralbf: Neural bilateral filtering for top-down instance segmentation on point clouds}.
\newblock In: \bibinfo{booktitle}{Proceedings of the IEEE/CVF Winter Conference on Applications of Computer Vision}, pp.~ \bibinfo{pages}{551--560}.  (\bibinfo{year}{2023})
\endplxcitation

\bibitem{wang2018sgpn}
\plxcitation{}{wang2018sgpn}{}{}{inproceedings}
\bibinfo{author}{\xsnm[Wang]\xfnm[, W.]}, et~al.: \bibinfo{title}{Sgpn: Similarity group proposal network for 3d point cloud instance segmentation}.
\newblock In: \bibinfo{booktitle}{Proceedings of the IEEE conference on computer vision and pattern recognition}, pp.~ \bibinfo{pages}{2569--2578}.  (\bibinfo{year}{2018})
\endplxcitation

\bibitem{wang2019associatively}
\plxcitation{}{wang2019associatively}{}{}{inproceedings}
\bibinfo{author}{\xsnm[Wang]\xfnm[, X.]}, et~al.: \bibinfo{title}{Associatively segmenting instances and semantics in point clouds}.
\newblock In: \bibinfo{booktitle}{Proceedings of the IEEE/CVF Conference on Computer Vision and Pattern Recognition}, pp.~ \bibinfo{pages}{4096--4105}.  (\bibinfo{year}{2019})
\endplxcitation

\bibitem{lahoud20193d}
\plxcitation{}{lahoud20193d}{}{}{inproceedings}
\bibinfo{author}{\xsnm[Lahoud]\xfnm[, J.]}, et~al.: \bibinfo{title}{3d instance segmentation via multi-task metric learning}.
\newblock In: \bibinfo{booktitle}{Proceedings of the IEEE/CVF International Conference on Computer Vision}, pp.~ \bibinfo{pages}{9256--9266}.  (\bibinfo{year}{2019})
\endplxcitation

\bibitem{pham2019jsis3d}
\plxcitation{}{pham2019jsis3d}{}{}{inproceedings}
\bibinfo{author}{\xsnm[Pham]\xfnm[, Q.H.]}, et~al.: \bibinfo{title}{Jsis3d: Joint semantic-instance segmentation of 3d point clouds with multi-task pointwise networks and multi-value conditional random fields}.
\newblock In: \bibinfo{booktitle}{Proceedings of the IEEE/CVF Conference on Computer Vision and Pattern Recognition}, pp.~ \bibinfo{pages}{8827--8836}.  (\bibinfo{year}{2019})
\endplxcitation

\bibitem{elich20193d}
\plxcitation{}{elich20193d}{}{}{inproceedings}
\bibinfo{author}{\xsnm[Elich]\xfnm[, C.]}, et~al.: \bibinfo{title}{3d bird’s-eye-view instance segmentation}.
\newblock In: \bibinfo{booktitle}{German Conference on Pattern Recognition}, pp.~ \bibinfo{pages}{48--61}. \bibinfo{organization}{Springer} (\bibinfo{year}{2019})
\endplxcitation

\bibitem{jiang2020pointgroup}
\plxcitation{}{jiang2020pointgroup}{}{}{inproceedings}
\bibinfo{author}{\xsnm[Jiang]\xfnm[, L.]}, et~al.: \bibinfo{title}{Pointgroup: Dual-set point grouping for 3d instance segmentation}.
\newblock In: \bibinfo{booktitle}{Proceedings of the IEEE/CVF Conference on Computer Vision and Pattern Recognition}, pp.~ \bibinfo{pages}{4867--4876}.  (\bibinfo{year}{2020})
\endplxcitation

\bibitem{han2020occuseg}
\plxcitation{}{han2020occuseg}{}{}{inproceedings}
\bibinfo{author}{\xsnm[Han]\xfnm[, L.]}, et~al.: \bibinfo{title}{Occuseg: Occupancy-aware 3d instance segmentation}.
\newblock In: \bibinfo{booktitle}{Proceedings of the IEEE/CVF conference on computer vision and pattern recognition}, pp.~ \bibinfo{pages}{2940--2949}.  (\bibinfo{year}{2020})
\endplxcitation

\bibitem{he2020instance}
\plxcitation{}{he2020instance}{}{}{inproceedings}
\bibinfo{author}{\xsnm[He]\xfnm[, T.]}, et~al.: \bibinfo{title}{Instance-aware embedding for point cloud instance segmentation}.
\newblock In: \bibinfo{booktitle}{European Conference on Computer Vision}, pp.~ \bibinfo{pages}{255--270}. \bibinfo{organization}{Springer} (\bibinfo{year}{2020})
\endplxcitation

\bibitem{zhao2020jsnet}
\plxcitation{}{zhao2020jsnet}{}{}{inproceedings}
\bibinfo{author}{\xsnm[Zhao]\xfnm[, L.]}, \bibinfo{author}{\xsnm[Tao]\xfnm[, W.]}: \bibinfo{title}{Jsnet: Joint instance and semantic segmentation of 3d point clouds}.
\newblock In: \bibinfo{booktitle}{Proceedings of the AAAI Conference on Artificial Intelligence}, vol.~\bibinfo{volume}{34}, pp.~ \bibinfo{pages}{12951--12958}.  (\bibinfo{year}{2020})
\endplxcitation

\bibitem{he2021dyco3d}
\plxcitation{}{he2021dyco3d}{}{}{inproceedings}
\bibinfo{author}{\xsnm[He]\xfnm[, T.]}, \bibinfo{author}{\xsnm[Shen]\xfnm[, C.]}, \bibinfo{author}{\xsnm[van~den Hengel]\xfnm[, A.]}: \bibinfo{title}{Dyco3d: Robust instance segmentation of 3d point clouds through dynamic convolution}.
\newblock In: \bibinfo{booktitle}{Proceedings of the IEEE/CVF Conference on Computer Vision and Pattern Recognition}, pp.~ \bibinfo{pages}{354--363}.  (\bibinfo{year}{2021})
\endplxcitation

\bibitem{chen2021hierarchical}
\plxcitation{}{chen2021hierarchical}{}{}{inproceedings}
\bibinfo{author}{\xsnm[Chen]\xfnm[, S.]}, et~al.: \bibinfo{title}{Hierarchical aggregation for 3d instance segmentation}.
\newblock In: \bibinfo{booktitle}{Proceedings of the IEEE/CVF International Conference on Computer Vision}, pp.~ \bibinfo{pages}{15467--15476}.  (\bibinfo{year}{2021})
\endplxcitation

\bibitem{zhang2021point}
\plxcitation{}{zhang2021point}{}{}{inproceedings}
\bibinfo{author}{\xsnm[Zhang]\xfnm[, B.]}, \bibinfo{author}{\xsnm[Wonka]\xfnm[, P.]}: \bibinfo{title}{Point cloud instance segmentation using probabilistic embeddings}.
\newblock In: \bibinfo{booktitle}{Proceedings of the IEEE/CVF Conference on Computer Vision and Pattern Recognition}, pp.~ \bibinfo{pages}{8883--8892}.  (\bibinfo{year}{2021})
\endplxcitation

\bibitem{chen2022jspnet}
\plxcitation{}{chen2022jspnet}{}{}{article}
\bibinfo{author}{\xsnm[Chen]\xfnm[, F.]}, et~al.: \bibinfo{title}{Jspnet: Learning joint semantic \& instance segmentation of point clouds via feature self-similarity and cross-task probability}.
\newblock \bibinfo{journal}{Pattern Recognition} \bibinfo{volume}{122}, \bibinfo{pages}{108250} (\bibinfo{year}{2022})
\endplxcitation

\bibitem{schult2023mask3d}
\plxcitation{}{schult2023mask3d}{}{}{inproceedings}
\bibinfo{author}{\xsnm[Schult]\xfnm[, J.]}, et~al.: \bibinfo{title}{Mask3d: Mask transformer for 3d semantic instance segmentation}.
\newblock In: \bibinfo{booktitle}{2023 IEEE International Conference on Robotics and Automation (ICRA)}, pp.~ \bibinfo{pages}{8216--8223}. \bibinfo{organization}{IEEE} (\bibinfo{year}{2023})
\endplxcitation

\bibitem{wu2019pointconv}
\plxcitation{}{wu2019pointconv}{}{}{inproceedings}
\bibinfo{author}{\xsnm[Wu]\xfnm[, W.]}, \bibinfo{author}{\xsnm[Qi]\xfnm[, Z.]}, \bibinfo{author}{\xsnm[Fuxin]\xfnm[, L.]}: \bibinfo{title}{Pointconv: Deep convolutional networks on 3d point clouds}.
\newblock In: \bibinfo{booktitle}{Proceedings of the IEEE/CVF Conference on Computer Vision and Pattern Recognition}, pp.~ \bibinfo{pages}{9621--9630}.  (\bibinfo{year}{2019})
\endplxcitation

\bibitem{qi2017pointnet++}
\plxcitation{}{qi2017pointnet++}{}{}{article}
\bibinfo{author}{\xsnm[Qi]\xfnm[, C.R.]}, et~al.: \bibinfo{title}{Pointnet++: Deep hierarchical feature learning on point sets in a metric space}.
\newblock \bibinfo{journal}{Advances in neural information processing systems} \bibinfo{volume}{30} (\bibinfo{year}{2017})
\endplxcitation

\bibitem{mo2019partnet}
\plxcitation{}{mo2019partnet}{}{}{inproceedings}
\bibinfo{author}{\xsnm[Mo]\xfnm[, K.]}, et~al.: \bibinfo{title}{Partnet: A large-scale benchmark for fine-grained and hierarchical part-level 3d object understanding}.
\newblock In: \bibinfo{booktitle}{Proceedings of the IEEE/CVF conference on computer vision and pattern recognition}, pp.~ \bibinfo{pages}{909--918}.  (\bibinfo{year}{2019})
\endplxcitation

\bibitem{armeni20163d}
\plxcitation{}{armeni20163d}{}{}{inproceedings}
\bibinfo{author}{\xsnm[Armeni]\xfnm[, I.]}, et~al.: \bibinfo{title}{3d semantic parsing of large-scale indoor spaces}.
\newblock In: \bibinfo{booktitle}{Proceedings of the IEEE conference on computer vision and pattern recognition}, pp.~ \bibinfo{pages}{1534--1543}.  (\bibinfo{year}{2016})
\endplxcitation

\bibitem{he2020learning}
\plxcitation{}{he2020learning}{}{}{inproceedings}
\bibinfo{author}{\xsnm[He]\xfnm[, T.]}, et~al.: \bibinfo{title}{Learning and memorizing representative prototypes for 3d point cloud semantic and instance segmentation}.
\newblock In: \bibinfo{booktitle}{European Conference on Computer Vision}, pp.~ \bibinfo{pages}{564--580}. \bibinfo{organization}{Springer} (\bibinfo{year}{2020})
\endplxcitation

\bibitem{denis2023improved}
\plxcitation{}{denis2023improved}{}{}{inproceedings}
\bibinfo{author}{\xsnm[Denis]\xfnm[, L.]}, \bibinfo{author}{\xsnm[Royen]\xfnm[, R.]}, \bibinfo{author}{\xsnm[Munteanu]\xfnm[, A.]}: \bibinfo{title}{Improved block merging for 3d point cloud instance segmentation}.
\newblock In: \bibinfo{booktitle}{2023 24th International Conference on Digital Signal Processing (DSP)}, pp.~ \bibinfo{pages}{1--5}. \bibinfo{organization}{IEEE} (\bibinfo{year}{2023})
\endplxcitation

\bibitem{vu2022softgroup}
\plxcitation{}{vu2022softgroup}{}{}{inproceedings}
\bibinfo{author}{\xsnm[Vu]\xfnm[, T.]}, et~al.: \bibinfo{title}{Softgroup for 3d instance segmentation on point clouds}.
\newblock In: \bibinfo{booktitle}{Proceedings of the IEEE/CVF Conference on Computer Vision and Pattern Recognition}, pp.~ \bibinfo{pages}{2708--2717}.  (\bibinfo{year}{2022})
\endplxcitation

\bibitem{qi2017pointnet}
\plxcitation{}{qi2017pointnet}{}{}{inproceedings}
\bibinfo{author}{\xsnm[Qi]\xfnm[, C.R.]}, et~al.: \bibinfo{title}{Pointnet: Deep learning on point sets for 3d classification and segmentation}.
\newblock In: \bibinfo{booktitle}{Proceedings of the IEEE conference on computer vision and pattern recognition}, pp.~ \bibinfo{pages}{652--660}.  (\bibinfo{year}{2017})
\endplxcitation

\bibitem{zhao2021point}
\plxcitation{}{zhao2021point}{}{}{inproceedings}
\bibinfo{author}{\xsnm[Zhao]\xfnm[, H.]}, et~al.: \bibinfo{title}{Point transformer}.
\newblock In: \bibinfo{booktitle}{Proceedings of the IEEE/CVF International Conference on Computer Vision}, pp.~ \bibinfo{pages}{16259--16268}.  (\bibinfo{year}{2021})
\endplxcitation

\bibitem{kingma2014adam}
\plxcitation{}{kingma2014adam}{}{}{article}
\bibinfo{author}{\xsnm[Kingma]\xfnm[, D.P.]}, \bibinfo{author}{\xsnm[Ba]\xfnm[, J.]}: \bibinfo{title}{Adam: A method for stochastic optimization}.
\newblock \bibinfo{journal}{arXiv preprint arXiv:1412.6980}  (\bibinfo{year}{2014})
\endplxcitation

\end{thebibliography}

\end{document}